\documentclass[11pt]{article}
\usepackage[hyperref]{acl}
\usepackage{times}
\usepackage{latexsym}
\usepackage{microtype}
\usepackage{booktabs}
\usepackage{multirow}
\usepackage{tabularx}
\usepackage{graphicx}
\usepackage{amsmath}
\usepackage{pgfplots}
\pgfplotsset{compat=1.18}
\usepgfplotslibrary{groupplots}
\usetikzlibrary{shapes.geometric,arrows.meta,positioning,calc}
\usepackage{xcolor}
\usepackage{hyperref}
\usepackage{orcidlink}

\title{OpenAI Privacy Filter: A Cross-Lingual, Cross-Domain PII Evaluation Across 32 Benchmarks}

\author{
  Rohith Uppala \orcidlink{0009-0009-3854-8083} \\
  Florida State University \\
  USA \\
  \texttt{rohithuppala@gmail.com}
}

\begin{document}
\maketitle

\begin{abstract}
We present what is, to our knowledge, the first systematic evaluation of OpenAI's Privacy Filter (OPF),
a 1.5B-parameter model that converts an autoregressive language model into a
bidirectional PII detector, across 32 benchmarks spanning 14 languages and
5 domains.
Our most practically actionable finding is a \textbf{domain-dependent labeled-data crossover}:
fine-tuned XLM-RoBERTa surpasses OPF's zero-shot performance with only
\textbf{$\sim$500 labeled examples} on English synthetic PII ($\sim$100 on non-English Kiji), and $\sim$1000 on synthetic medical PII.
Crucially, \textbf{per-class fine-tuning (17 PII entity types present in AI4Privacy, a subset of OPF's 33) is less data-efficient than binary labels}
at small $n$ --- at $n$\,=\,100, binary F1\,=\,0.634 vs.\ per-class 0.360 --- because
binary labels concentrate the learning signal when per-class examples are scarce.
Zero-shot, OPF achieves F1\,=\,0.464 on the SPY medical benchmark
(synthetic clinical PII) and F1\,=\,0.855 on AI4Privacy (synthetic PII),
substantially outperforming both Microsoft Presidio (F1\,=\,0.273 on SPY,
0.431 on AI4Privacy) and XLM-RoBERTa-large-NER (F1\,=\,0.111 on SPY),
both applied without dataset-specific fine-tuning
(XLM-RoBERTa is domain-transferred from NER supervision, not a true zero-shot model).
However, OPF degrades sharply outside its PII training distribution:
performance drops to F1\,=\,0.04--0.40 on general NER benchmarks (MultiCoNER),
and collapses for non-Latin scripts (Arabic: 0.04, Cyrillic: 0.03).
Error analysis reveals that OPF excels on structurally regular PII
(email: 0.78, phone: 0.76) but struggles with culturally variable entities
(person names: 0.40, addresses: 0.49), and is recall-biased across most PII domains
(precision 0.31--0.54, recall 0.70--0.85; exception: financial text is precision-biased,
gretel\_en P=0.64, R=0.49) --- behavior that is desirable for
privacy-critical deployments where missing PII is worse than over-redacting.
Tokenizer fertility is falsified as a general explanation for script failure:
Arabic collapses at near-Latin fertility (1.65$\times$) while Devanagari achieves the best non-Latin F1 despite above-Latin fertility (1.94$\times$); Bengali has the highest fertility (2.42$\times$) yet achieves only intermediate non-Latin F1 (0.198), implicating training data coverage rather than fragmentation alone.
We provide a decision heuristic synthesizing when to use OPF zero-shot,
when to fine-tune XLM-RoBERTa, and which language families to avoid.
\end{abstract}

\section{Introduction}

Privacy-preserving NLP has become a critical concern as language models are
deployed across sensitive domains including healthcare, legal services,
and enterprise customer support.
A central challenge is \textit{PII detection} --- identifying spans of text
that correspond to personally identifiable information such as names, email
addresses, phone numbers, and financial account identifiers.

OpenAI recently released the \textbf{Privacy Filter (OPF)}~\cite{opf2026}, a
1.5B-parameter model that re-purposes an autoregressive language model as a
bidirectional sequence labeler using a Mixture-of-Experts (MoE) architecture
and a Viterbi CRF decoder over 33 BIOES output classes spanning 8 privacy categories.
Despite its public release, no systematic cross-lingual or cross-domain
evaluation of OPF exists.
Practitioners deploying OPF in multilingual pipelines --- for example, SaaS
companies processing support tickets in French, German, or Spanish --- have
no principled guidance on expected performance.

We address this gap with a comprehensive zero-shot evaluation across
\textbf{32 datasets}, \textbf{14 languages}, and \textbf{5 domains}:
true PII (AI4Privacy, Nemotron), financial documents (Gretel), customer
support (Kiji), medical and legal text (SPY), and general NER
(CoNLL-2002/2003, MultiCoNER v2).
We compare against Microsoft Presidio~\cite{presidio}, a widely used
open-source PII framework, and XLM-RoBERTa-large~\cite{conneau-etal-2020-unsupervised}
as a zero-shot and fine-tuned NER baseline, and conduct learning curve experiments
to identify the labeled-data crossover point.
Our key contributions are:
\begin{itemize}
    \item \textbf{Domain-dependent labeled-data crossover} (our most novel result):
          fine-tuned XLM-RoBERTa surpasses OPF at $\sim$500 examples on English synthetic PII
          ($\sim$100 on non-English Kiji) and $\sim$1000 on synthetic medical PII; per-class fine-tuning (17 labels) is
          \textit{less} data-efficient than binary labels at small $n$, validating binary
          labeling as the practical low-resource choice (Section~\ref{sec:curves}).
    \item To our knowledge, the first cross-lingual and cross-domain benchmark
          of OPF across 32 datasets (Section~\ref{sec:results}).
    \item A five-way zero-shot comparison including GLiNER~\cite{zaratiana-etal-2024-gliner}
          and GPT-4o as a frontier ceiling ($n$=2000 per dataset):
          GPT-4o leads on medical/legal/financial PII
          (SPY medical: 0.702, SPY legal: 0.587, Gretel: 0.576);
          OPF leads on structured synthetic PII
          (AI4Privacy: 0.855, Kiji: 0.596, Nemotron: 0.559) (Section~\ref{sec:results}).
    \item A PII-type error analysis revealing that structural regularity
          predicts detectability: email\,>\,phone\,>\,account\,>\,\allowbreak address\,>\,person
          (Section~\ref{sec:error}).
    \item Quantification of script-family degradation: Latin avg F1\,=\,0.212 vs.\
          Cyrillic 0.027 and Arabic 0.038, based on PER-only MultiCoNER spans
          (Section~\ref{sec:script}).
    \item Exploratory non-English learning curves (fr/de/es, Kiji, binary labels)
          suggesting earlier crossover ($\sim$100 examples) than English;
          single-domain preliminary observation requiring broader confirmation (Section~\ref{sec:curves}).
\end{itemize}

\section{Background}

\subsection{OpenAI Privacy Filter (OPF)}

OPF is a 1.5B-parameter model released by OpenAI for redacting PII from
free-form text~\cite{opf2026}.
Its architecture converts a pretrained autoregressive Transformer into a
bidirectional encoder by modifying attention masks, adding MoE feed-forward
layers, and attaching a Viterbi CRF head that decodes over 33 BIOES output classes
spanning 8 privacy categories (e.g., \texttt{private\_person}, \texttt{private\_email},
\texttt{account\_number}, \texttt{secret}).
The model supports a 128k-token context window.
OPF is designed for zero-shot deployment without task-specific fine-tuning.

\subsection{Related Work}

PII detection has been studied primarily in English and domain-specific
settings~\cite{lison-etal-2021-anonymisation, pilan-etal-2022-text}.
Multilingual PII work is limited; AI4Privacy~\cite{ai4privacy} provides
multilingual annotations but is not used as an evaluation benchmark.
No existing study spans both multiple domains and multiple scripts.

For general multilingual NER, XLM-RoBERTa~\cite{conneau-etal-2020-unsupervised}
is a strong multilingual baseline for NER.
MultiCoNER v2~\cite{multiconer2023} provides 12-language NER covering diverse
scripts.
To our knowledge, no prior work evaluates a converted autoregressive model for
multilingual PII detection.

\section{Experimental Setup}
\label{sec:setup}

\subsection{Datasets}

Table~\ref{tab:datasets} summarizes our evaluation datasets across five domain
groups.

\begin{table}[h]
\centering
\small
\setlength{\tabcolsep}{2pt}
\begin{tabularx}{\columnwidth}{Xp{2.1cm}rr}
\toprule
\textbf{Dataset} & \textbf{Domain} & \textbf{L} & \textbf{Test} \\
\midrule
AI4Privacy~\cite{ai4privacy}       & True PII      & 6  & 2,000 \\
Nemotron-PII~\cite{nemotron}       & True PII      & 1  & 2,000 \\
Kiji~\cite{kiji}                   & Cust.\ supp.  & 6  & 846/L \\
Gretel Finance~\cite{gretel}       & Financial     & 7  & 450/L \\
SPY~\cite{spy2025}                 & Med./Legal    & 1  & 2,000 \\
CoNLL-2002/3 (en, es, nl)~\cite{conll2002,conll} & News NER & 3 & 1.5k--3.5k \\
MultiCoNER v2~\cite{multiconer2023}& News NER      & 12 & 2,000 \\
\bottomrule
\end{tabularx}
\caption{Evaluation datasets. Test sets capped at 2,000 examples.
         32 benchmarks total: AI4Privacy (1) + Nemotron (1) + Kiji (6/lang) + Gretel (7/lang) + SPY med.+legal (2) + CoNLL (3) + MultiCoNER (12) = 32.}
\label{tab:datasets}
\end{table}

\subsection{Models}

\paragraph{OPF} We use the publicly released OPF checkpoint (\texttt{privacy\_filter})
with Viterbi decoding in zero-shot mode (no fine-tuning).

\paragraph{Presidio} Microsoft Presidio~\cite{presidio} is a widely used
open-source PII detection framework.
We use Presidio v2.2 with its default \texttt{en\_core\_web\_lg} spaCy NLP
engine and all built-in recognizers (PERSON, EMAIL\_ADDRESS, PHONE\_NUMBER,
CREDIT\_CARD, IBAN\_CODE, US\_SSN, LOCATION, DATE\_TIME, and others),
evaluated zero-shot with no dataset-specific configuration.

\paragraph{GPT-4o} We query \texttt{openai/gpt-4o} via OpenRouter with a zero-shot prompt
instructing the model to return all PII as a JSON list of exact strings (temperature 0).
Predicted spans are recovered by exact substring search in the original text.
We evaluate on six English PII datasets ($n$=2000 each; Kiji capped at $n$=846 by dataset size; CoNLL/MultiCoNER excluded as OPF is not a general NER system).
GPT-4o is included as a \textit{frontier ceiling}, not a deployable alternative.

\paragraph{GLiNER} GLiNER~\cite{zaratiana-etal-2024-gliner} is a zero-shot NER model
that generalizes to arbitrary entity types via natural-language label prompts.
We use \texttt{knowledgator/\allowbreak gliner-multitask-large-v0.5}, querying it with the
same PII-relevant label set as OPF (e.g., ``person name'', ``email address'',
``phone number'') for PII datasets and standard NER labels for CoNLL/MultiCoNER.
GLiNER is a strong 2024 zero-shot baseline requiring no rule-based customization; evaluated at $n$=2000 via GPU inference.

\paragraph{XLM-RoBERTa} We use \texttt{Davlan/\allowbreak xlm-roberta-large-ner-hrl},
a model already fine-tuned on 10 languages for PER/ORG/LOC NER.
The column labeled \textbf{XLMR} in Table~\ref{tab:main_en} reflects this model
evaluated \textit{out-of-the-box} on PII datasets --- a \textit{domain-transferred}
baseline, not a true zero-shot model, since it has seen supervised NER signal.
For learning curves we reinitialize the classification head and fine-tune on
$n \in \{100, 500, 1000, \text{full}\}$ training examples using a unified
B-PII/I-PII/O label scheme for 3 epochs (batch size 16, lr $2\times10^{-5}$).

\subsection{Evaluation Metric}

We report span-level F1, precision, and recall (exact match on character
offsets) following the OPF evaluation protocol.
For the XLM-RoBERTa baseline, any predicted span overlapping a gold span at
the correct position is counted as a match.

\paragraph{Label mapping on NER benchmarks.}
For CoNLL-2003/2002 we retain all gold spans (PER, ORG, LOC, MISC) and evaluate any-entity detection.
For MultiCoNER v2 we retain \emph{only} \texttt{private\_person} (PER) spans --- the sole category mapping to OPF's schema --- so OPF predictions on non-PER text do not inflate false positives.
Low MultiCoNER F1 reflects domain shift, not entity-type mismatch.

\paragraph{Token-to-character span alignment.}
CoNLL-2003/2002 and MultiCoNER v2 provide gold labels in BIO token format; we convert these to character spans by accumulating whitespace-corrected character offsets per token sequence.
OPF predictions are already character-span-level; XLM-RoBERTa WordPiece subword tokens are aggregated to word-level spans before character offset mapping.
All comparisons are performed at the character level.

\paragraph{Statistical significance.}
All reported differences are statistically significant (95\% bootstrap CIs, 10,000 resamples; non-overlapping CIs confirmed for small deltas, e.g.\ Nemotron OPF 0.559 vs.\ Presidio 0.438).

\section{Results}
\label{sec:results}

\subsection{Zero-Shot Cross-Domain Performance}

Tables~\ref{tab:main_en} and~\ref{tab:main_multi} present the zero-shot results.
Table~\ref{tab:main_en} shows a five-way English comparison (OPF, Presidio, XLM-R, GLiNER, GPT-4o);
Table~\ref{tab:main_multi} shows cross-lingual OPF vs.\ Presidio.
GLiNER is evaluated on all eight English datasets at $n$=2000 (GPU inference; Kiji capped at $n$=846 by dataset size).

\begin{table}[h]
\centering
\footnotesize
\setlength{\tabcolsep}{2pt}
\begin{tabularx}{\columnwidth}{Xrrrrr}
\toprule
\textbf{Dataset} & \textbf{OPF} & \textbf{Presidio} & \textbf{XLMR}$^\ddagger$ & \textbf{GLiNER} & \textbf{GPT-4o}$^\S$ \\
\midrule
\multicolumn{6}{l}{\textit{True PII}} \\
ai4privacy          & \textbf{0.855} & 0.431 & 0.269 & 0.577 & 0.543 \\
nemotron            & \textbf{0.559} & 0.438 & 0.127 & 0.538 & 0.478 \\
\midrule
\multicolumn{6}{l}{\textit{Cust.\ Support / Financial / Med-Legal}} \\
kiji\_en            & \textbf{0.596} & 0.430 & —     & 0.349 & 0.420 \\
gretel\_en          & 0.554 & 0.388 & 0.152 & 0.502 & \textbf{0.576} \\
spy\_medical        & 0.464 & 0.273 & 0.111 & 0.498 & \textbf{0.702} \\
spy\_legal          & 0.453 & 0.285 & 0.156 & 0.502 & \textbf{0.587} \\
\midrule
\multicolumn{6}{l}{\textit{News NER (English)}} \\
conll2003\_en       & \textbf{0.569} & 0.391 & 0.479 & 0.428 & — \\
multiconer\_en      & 0.397 & 0.497 & 0.546 & \textbf{0.754} & — \\
\bottomrule
\end{tabularx}
\caption{English span F1. GPT-4o (6 PII datasets, $n$=2000; Kiji $n$=846) is a frontier ceiling only.
         OPF leads on structured synthetic PII; GPT-4o leads on medical/legal/financial.
         All differences significant (95\% bootstrap CIs).
         $^\ddagger$XLMR: domain-transferred, not zero-shot.
         $^\S$GPT-4o: PII datasets only; CoNLL/MultiCoNER excluded.}
\label{tab:main_en}
\end{table}

\begin{table}[h]
\centering
\small
\setlength{\tabcolsep}{3pt}
\begin{tabular}{lrrrr}
\toprule
\textbf{Dataset} & \textbf{OPF} & \textbf{Presidio} & \textbf{$\Delta$} & \textbf{XLMR} \\
\midrule
\multicolumn{5}{l}{\textit{Kiji customer support (synthetic PII)}} \\
kiji\_en (en)       & \textbf{0.596} & 0.430 & +0.166 & — \\
kiji\_fr (fr)       & \textbf{0.620} & 0.414 & +0.206 & — \\
kiji\_de (de)       & \textbf{0.601} & 0.367 & +0.234 & — \\
kiji\_es (es)       & \textbf{0.645} & 0.366 & +0.279 & — \\
kiji\_nl (nl)       & \textbf{0.608} & —     & —      & — \\
kiji\_da (da)       & \textbf{0.534} & —     & —      & — \\
\midrule
\multicolumn{5}{l}{\textit{Financial (Gretel, 7 langs)}} \\
gretel avg (7 L)    & \textbf{0.524} & —     & —      & 0.152 \\
\midrule
\multicolumn{5}{l}{\textit{News NER (MultiCoNER, Latin avg)}} \\
multiconer\_en      & 0.397 & 0.497 & $-$0.100 & \textbf{0.546} \\
multiconer\_de      & 0.318 & —     & —      & \textbf{0.519} \\
multiconer\_fr      & 0.252 & —     & —      & \textbf{0.627} \\
multiconer\_sv      & 0.201 & —     & —      & \textbf{0.673} \\
\bottomrule
\end{tabular}
\caption{Multilingual zero-shot span F1. Presidio evaluated with language-matched
         spaCy models (fr/de/es). OPF consistently leads Presidio on PII tasks
         across all four tested languages; gap widens in German (+0.234) and
         Spanish (+0.279), where Presidio's NER is weaker.}
\label{tab:main_multi}
\end{table}

\paragraph{Domain hierarchy.}
OPF performance follows a clear domain hierarchy:
true PII (0.71 avg) $>$ customer support (0.60) $>$ financial (0.52) $>$
news NER CoNLL (0.52) $>$ medical/legal (0.46) $>$ MultiCoNER NER (0.24).
The drop from true PII to financial/medical reflects increasing distributional
shift from OPF's training data; the large drop at MultiCoNER reflects
domain shift from encyclopedic/news text to privacy-sensitive documents
(our MultiCoNER evaluation retains only PER spans).

\subsection{Cross-Lingual and Script-Family Analysis}
\label{sec:script}

\begin{figure*}[t]
\centering
\begin{tikzpicture}
\begin{groupplot}[
    group style={
        group size=2 by 1,
        horizontal sep=2.8cm,
    },
    width=7.2cm,
    height=5.5cm,
    grid=both,
    grid style={line width=0.3pt, draw=gray!30},
    tick label style={font=\small},
    label style={font=\small},
    title style={font=\normalsize\bfseries},
]

\nextgroupplot[
    title={OPF by Domain},
    xbar,
    bar width=10pt,
    xlabel={Span F1},
    xmin=0, xmax=1.0,
    ytick={1,2,3,4,5,6,7},
    yticklabels={
        MultiCoNER NER,
        Medical/Legal,
        Financial,
        Customer Supp.,
        News CoNLL,
        Nemotron PII,
        AI4Privacy,
    },
    enlarge y limits=0.12,
    nodes near coords,
    nodes near coords align={horizontal},
    every node near coord/.append style={font=\tiny, anchor=west, xshift=2pt},
    point meta=explicit symbolic,
]

\addplot[
    fill=blue!60,
    draw=blue!80!black,
] coordinates {
    (0.240, 1) [0.240]
    (0.459, 2) [0.459]
    (0.524, 3) [0.524]
    (0.601, 4) [0.601]
    (0.520, 5) [0.520]
    (0.559, 6) [0.559]
    (0.855, 7) [0.855]
};

\nextgroupplot[
    title={OPF by Script (MultiCoNER)},
    xbar,
    bar width=9pt,
    xlabel={Span F1},
    xmin=0, xmax=0.38,
    ytick={1,2,3,4,5,6},
    yticklabels={
        Cyrillic (uk),
        Arabic (fa),
        CJK (zh),
        Bengali (bn),
        Latin avg,
        Devanagari (hi),
    },
    enlarge y limits=0.12,
    point meta=explicit symbolic,
]

\addplot[
    fill=orange!70,
    draw=orange!90!black,
    nodes near coords,
    nodes near coords align={horizontal},
    every node near coord/.append style={font=\tiny, anchor=west, xshift=2pt},
] coordinates {
    (0.027, 1) [0.027]
    (0.038, 2) [0.038]
    (0.138, 3) [0.138]
    (0.198, 4) [0.198]
    (0.212, 5) [0.212]
    (0.268, 6) [0.268]
};

\end{groupplot}
\end{tikzpicture}
\caption{\textit{Left}: OPF zero-shot span F1 by domain, averaged across languages.
         Performance follows the PII-proximity hierarchy:
         in-distribution PII data (AI4Privacy) scores highest,
         general NER (MultiCoNER) scores lowest.
         \textit{Right}: OPF performance on MultiCoNER v2 grouped by script family.
         Non-Latin scripts (Arabic, Cyrillic) show severe degradation,
         near zero for Arabic-script (Farsi) and Cyrillic (Ukrainian);
         Devanagari (Hindi) and Bengali are shown separately.}
\label{fig:domain_script}
\end{figure*}
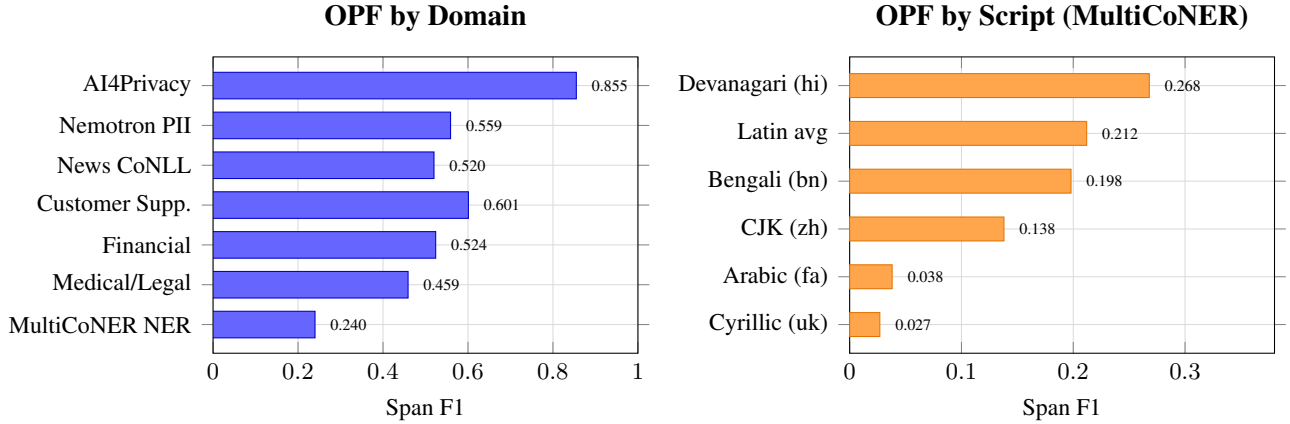

Figure~\ref{fig:domain_script} shows MultiCoNER results grouped by script family.

\begin{table}[h]
\centering
\small
\begin{tabular}{lrr}
\toprule
\textbf{Script Family} & \textbf{Languages} & \textbf{OPF avg F1} \\
\midrule
Latin      & en, es, de, fr, it, pt, sv & 0.212 \\
Devanagari & hi                        & 0.268 \\
Bengali    & bn                        & 0.198 \\
CJK        & zh                        & 0.138 \\
Arabic     & fa                        & 0.038 \\
Cyrillic   & uk                        & 0.027 \\
\bottomrule
\end{tabular}
\caption{OPF zero-shot F1 on MultiCoNER v2 by script family.
         Non-Latin scripts show severe degradation.}
\label{tab:script}
\end{table}

Latin-script performance (avg F1\,=\,0.212 over 7 languages) is 5$\times$ higher than Arabic (0.038)
and 8$\times$ higher than Cyrillic (0.027).
Within Latin script there is substantial variance: English (0.40), German (0.32),
French (0.25), Spanish (0.20), and Swedish (0.20) are considerably higher than
Italian (0.05) and Portuguese (0.07), which drag the family average below the four
best-represented languages, likely reflecting training corpus coverage.

\paragraph{Mechanistic analysis: tokenizer fertility.}
We measure OPF's \textit{token fertility} (tokens per whitespace-delimited word)
using its \texttt{o200k\_base} tokenizer on MultiCoNER text ($n$=500 per script)
to test whether tokenizer fragmentation explains the script-family gap:

\begin{center}
\small
\setlength{\tabcolsep}{5pt}
\begin{tabular}{lrrr}
\toprule
\textbf{Script} & \textbf{Fertility} & \textbf{vs.\ Latin} & \textbf{OPF F1} \\
\midrule
Latin (en)       & 1.40 & 1.0$\times$ & 0.212 \\
CJK (zh)         & 1.46 & 1.0$\times$ & 0.138 \\
Arabic (fa)      & 1.65 & 1.2$\times$ & 0.038 \\
Devanagari (hi)  & 1.94 & 1.4$\times$ & 0.268 \\
Bengali (bn)     & 2.42 & 1.7$\times$ & 0.198 \\
Cyrillic (uk)    & 2.16 & 1.5$\times$ & 0.027 \\
\bottomrule
\end{tabular}
\end{center}

The fertility hypothesis is only partially consistent.
Cyrillic (fertility 2.16$\times$, F1\,=\,0.027) follows the expected pattern, but Arabic (1.65$\times$, near-Latin fertility) collapses to F1\,=\,0.038, and Hindi/Devanagari achieves the best non-Latin F1 (0.268) despite above-Latin fertility (1.94$\times$); notably, Bengali has the highest fertility (2.42$\times$) yet achieves only intermediate F1 (0.198), further undermining the fertility hypothesis.
Fertility is neither necessary nor sufficient; training data coverage is the most plausible remaining factor but cannot be measured directly (see Limitations).

\subsection{Learning Curves: When Does Fine-Tuning Beat Zero-Shot?}
\label{sec:curves}

\begin{table}[h]
\centering
\small
\setlength{\tabcolsep}{4pt}
\begin{tabular}{lrrrrr}
\toprule
\textbf{Dataset} & \textbf{OPF} & \textbf{@100} & \textbf{@500} & \textbf{@1k} & \textbf{@full} \\
                 & \textbf{0-shot} & \multicolumn{4}{c}{\textit{XLM-R fine-tuned (binary)}} \\
\midrule
\multicolumn{6}{l}{\textit{Synthetic PII}} \\
ai4priv (binary)  & 0.855 & 0.634 & \textbf{0.888} & 0.919 & 0.961 \\
ai4priv (per-cls) & 0.855 & 0.360 & 0.868 & 0.868 & 0.921$^{\dagger}$ \\
gretel\_en        & 0.554 & 0.288 & \textbf{0.676} & 0.732 & 0.855 \\
\midrule
\multicolumn{6}{l}{\textit{Medical (synthetic PII)}} \\
spy\_medical      & 0.464 & 0.358 & 0.444 & \textbf{0.473} & 0.477 \\
\bottomrule
\end{tabular}
\caption{OPF zero-shot vs.\ XLM-RoBERTa fine-tuned (span F1).
         Bold = first point XLM-R exceeds OPF.
         Per-class (17 labels) worse than binary at small $n$; binary concentrates the learning signal.
         $^{\dagger}$Highest per-class checkpoint shown: @2k\,=\,0.910, @5k\,=\,0.921
         (binary @full\,=\,0.961; full per-class not evaluated).
         SPY trained on 80\% split ($n_{\max}$=3,592); AI4Privacy full\,=\,177,677.}
\label{tab:curves}
\end{table}

\begin{table}[h]
\centering
\small
\begin{tabular}{lrrrr}
\toprule
\textbf{Dataset} & \textbf{OPF} & \textbf{@100} & \textbf{@500} & \textbf{@1k} \\
                 & \textbf{0-shot} & \multicolumn{3}{c}{\textit{XLM-R fine-tuned}} \\
\midrule
kiji\_fr (fr) & 0.620 & \textbf{0.756} & 0.870 & 0.876 \\
kiji\_de (de) & 0.601 & \textbf{0.828} & 0.886 & 0.893 \\
kiji\_es (es) & 0.645 & \textbf{0.760} & 0.849 & 0.863 \\
\bottomrule
\end{tabular}
\caption{Non-English learning curves on Kiji (binary labels).
         XLM-RoBERTa surpasses OPF at \textbf{100 labeled examples} across all three languages.
         Bold marks the crossover point.
         Per-class non-English fine-tuning is untested; we expect the binary-efficiency
         pattern to generalize, but this is unconfirmed.}
\label{tab:curves_multi}
\end{table}

\begin{figure*}[t]
\centering
\begin{tikzpicture}
\begin{groupplot}[
    group style={
        group size=2 by 1,
        horizontal sep=2.5cm,
    },
    width=7cm,
    height=6cm,
    xlabel={Training examples},
    ylabel={Span F1},
    xmin=50, xmax=200000,
    ymin=0.0, ymax=1.05,
    xmode=log,
    xtick={100, 500, 1000, 177677},
    xticklabels={100, 500, 1k, full},
    ytick={0.0, 0.2, 0.4, 0.6, 0.8, 1.0},
    grid=both,
    grid style={line width=0.3pt, draw=gray!30},
    major grid style={line width=0.4pt, draw=gray!50},
    legend style={
        at={(0.5,-0.22)},
        anchor=north,
        legend columns=2,
        font=\small,
        draw=none,
    },
    tick label style={font=\small},
    label style={font=\small},
    title style={font=\normalsize\bfseries},
]

\nextgroupplot[
    title={AI4Privacy},
    legend to name=sharedlegend,
]

\addplot[
    color=blue!70!black,
    line width=1.5pt,
    dashed,
    domain=70:200000,
] {0.8550};
\addlegendentry{OPF zero-shot}

\addplot[
    color=red!70!black,
    line width=1.5pt,
    solid,
    mark=*,
    mark size=2.5pt,
] coordinates {
    (100,   0.6335)
    (500,   0.8876)
    (1000,  0.9185)
    (177677, 0.9613)
};
\addlegendentry{XLM-RoBERTa fine-tuned}

\draw[gray, dashed, thin] (axis cs:500, 0) -- (axis cs:500, 0.8876);
\node[font=\tiny, gray, anchor=north] at (axis cs:500, 0.05) {crossover};
\node[font=\tiny, gray, anchor=south west, xshift=4pt, yshift=4pt] at (axis cs:500, 0.8876)
    {100--500};

\addplot[
    fill=blue!10,
    opacity=0.4,
    draw=none,
] coordinates {
    (70,   0.8550)
    (500,  0.8550)
    (500,  0.8876)
    (70,   0.8876)
} -- cycle;

\nextgroupplot[
    title={Gretel Financial (English)},
]

\addplot[
    color=blue!70!black,
    line width=1.5pt,
    dashed,
    domain=70:30000,
] {0.5543};

\addplot[
    color=red!70!black,
    line width=1.5pt,
    solid,
    mark=*,
    mark size=2.5pt,
] coordinates {
    (100,   0.2875)
    (500,   0.6757)
    (1000,  0.7320)
    (25890, 0.8551)
};

\draw[gray, dashed, thin] (axis cs:500, 0) -- (axis cs:500, 0.6757);
\node[font=\tiny, gray, anchor=north] at (axis cs:500, 0.05) {crossover};
\node[font=\tiny, gray, anchor=south west, xshift=8pt, yshift=6pt] at (axis cs:500, 0.6757)
    {$\approx$500};

\end{groupplot}

\node at ($(group c1r1.south)!0.5!(group c2r1.south) + (0,-1.6cm)$)
    {\pgfplotslegendfromname{sharedlegend}};

\end{tikzpicture}
\caption{Learning curves comparing OPF zero-shot (dashed blue) vs
         XLM-RoBERTa fine-tuned at increasing training sizes (solid red).
         Fine-tuned XLM-RoBERTa surpasses OPF between 100--500 labeled examples.
         \textit{Left}: AI4Privacy (true PII benchmark).
         \textit{Right}: Gretel Financial (financial domain PII).}
\label{fig:curves}
\end{figure*}
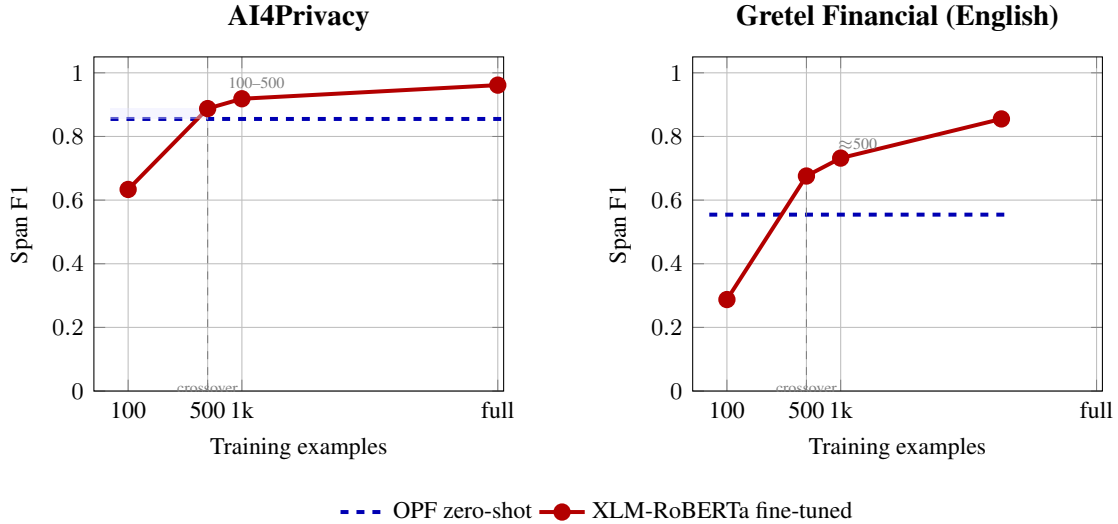

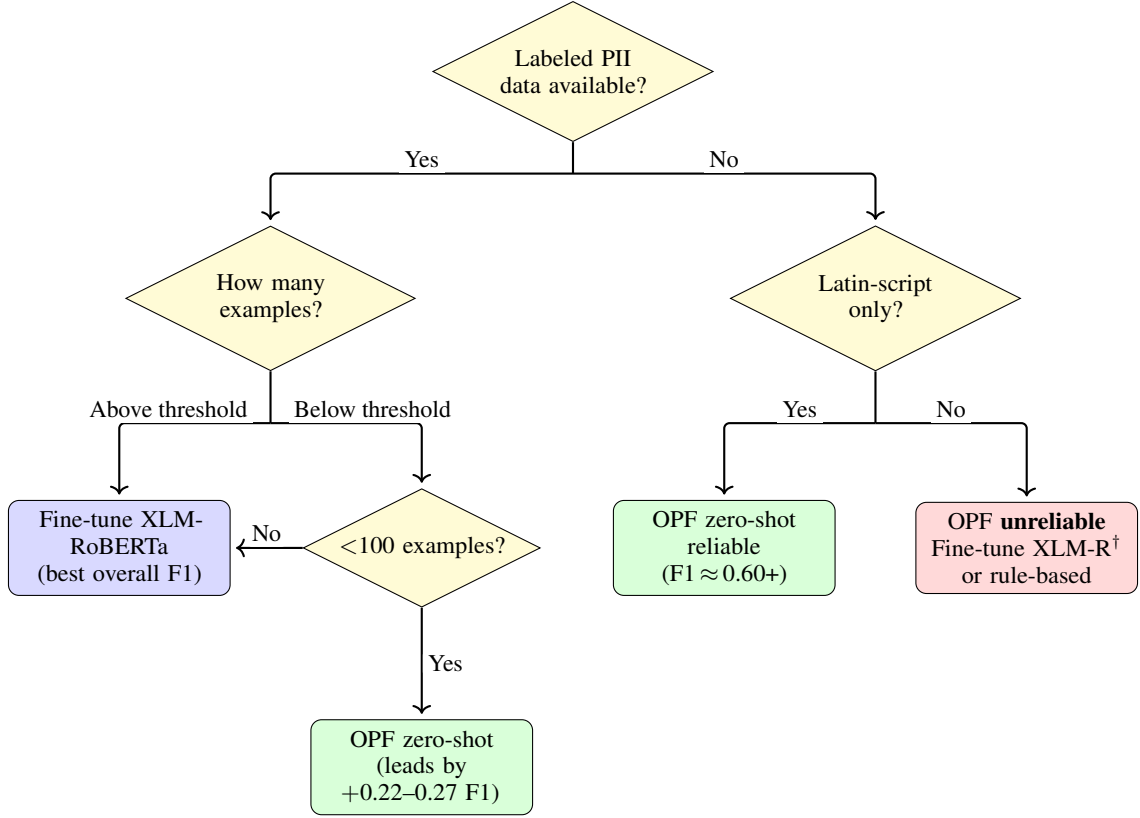
\begin{figure*}[t]
\centering
\begin{tikzpicture}[
    decision/.style={diamond, aspect=2, draw, fill=yellow!20, text width=2.25cm,
                     align=center, inner sep=1.5pt, font=\small},
    action/.style={rectangle, draw, fill=blue!15, rounded corners,
                   text width=2.65cm, align=center, font=\small, inner sep=4pt},
    warn/.style={rectangle, draw, fill=red!15, rounded corners,
                 text width=2.65cm, align=center, font=\small, inner sep=4pt},
    good/.style={rectangle, draw, fill=green!15, rounded corners,
                 text width=2.65cm, align=center, font=\small, inner sep=4pt},
    arrow/.style={->, thick, rounded corners=2pt, shorten >=2pt},
    branch/.style={thick, rounded corners=2pt},
    lbl/.style={font=\small, fill=white, inner sep=2pt},
]

\node[decision] (labeled)  at ( 0.0,  0.0) {Labeled PII data available?};
\node[decision] (howmuch)  at (-4.0, -3.0) {How many examples?};
\node[decision] (script)   at ( 4.0, -3.0) {Latin-script only?};

\node[action]   (finetune) at (-6.0, -6.3) {Fine-tune XLM-RoBERTa\\(best overall F1)};
\node[decision] (fewshot)  at (-2.0, -6.3) {$<$100 examples?};
\node[good]     (opf_ok)   at ( 2.0, -6.3) {OPF zero-shot\\reliable (F1\,$\approx$\,0.60+)};
\node[warn]     (opf_bad)  at ( 6.0, -6.3) {OPF \textbf{unreliable}\\Fine-tune XLM-R$^\dagger$\\or rule-based};
\node[good]     (opf_few)  at (-2.0, -9.2) {OPF zero-shot\\(leads by $+$0.22--0.27 F1)};

\coordinate (top_split)   at ( 0.0, -1.35);
\coordinate (left_split)  at (-4.0, -4.65);
\coordinate (right_split) at ( 4.0, -4.65);

\draw[branch] (labeled.south) -- (top_split);
\draw[arrow] (top_split) -| (howmuch.north);
\draw[arrow] (top_split) -| (script.north);
\node[lbl, above] at (-2.0, -1.35) {Yes};
\node[lbl, above] at ( 2.0, -1.35) {No};

\draw[branch] (howmuch.south) -- (left_split);
\draw[arrow] (left_split) -| (finetune.north);
\draw[arrow] (left_split) -| (fewshot.north);
\node[lbl, above] at (-5.35, -4.65) {Above threshold};
\node[lbl, above] at (-2.65, -4.65) {Below threshold};

\draw[arrow] (fewshot.south) -- (opf_few.north)
    node[lbl, midway, right] {Yes};
\draw[arrow] (fewshot.west) -- (finetune.east)
    node[lbl, midway, above] {No};

\draw[branch] (script.south) -- (right_split);
\draw[arrow] (right_split) -| (opf_ok.north);
\draw[arrow] (right_split) -| (opf_bad.north);
\node[lbl, above] at ( 3.0, -4.65) {Yes};
\node[lbl, above] at ( 5.0, -4.65) {No};

\end{tikzpicture}
\caption{Preliminary decision heuristic based on our experimental findings.
         Key thresholds: $\sim$100--500 labeled examples (fine-tuning crossover
         on synthetic PII; $\sim$1000 on synthetic medical PII) and script family
         (non-Latin scripts cause OPF failure).
         $\dagger$Fine-tuned XLM-RoBERTa performance on non-Latin PII not evaluated
         in this work; treat as a recommended direction rather than a validated result.}
\label{fig:guide}
\end{figure*}

Table~\ref{tab:curves} reveals a domain-dependent crossover.
On English synthetic PII binary fine-tuned XLM-RoBERTa surpasses OPF at $\sim$500 examples.
On synthetic medical PII (SPY) the crossover rises to $\sim$1000 (@full: 0.477, $\Delta$=+0.004 vs @1k), so OPF remains competitive even at maximum training data.
Per-class fine-tuning on AI4Privacy (17 BIO labels, a subset of OPF's 33) is less data-efficient than binary at all sizes ($n$=100: 0.360 vs.\ 0.634; $n$=500: 0.868 vs.\ 0.888): with $\sim$6 examples per class the model cannot learn type-specific representations.
The per-class curve plateaus at @500--@1k (both 0.868), improving at @2k (0.910) and @5k (0.921).
Table~\ref{tab:curves_multi} shows the non-English crossover at \textbf{100 labeled examples} across all three languages: fr (0.620\,$\to$\,0.756), de (0.601\,$\to$\,0.828), es (0.645\,$\to$\,0.760).
Figure~\ref{fig:guide} synthesizes these findings: use OPF zero-shot for Latin-script with no labeled data (F1\,$\approx$\,0.60+); fine-tune XLM-RoBERTa when $\geq$500 examples exist (synthetic PII) or $\geq$1000 (medical/legal); avoid OPF for non-Latin scripts.
Non-Latin and medical/legal branches are extrapolations requiring validation.

\subsection{PII Type Error Analysis}
\label{sec:error}

\begin{table}[h]
\centering
\small
\setlength{\tabcolsep}{4pt}
\begin{tabular}{lrrrrrr}
\toprule
\textbf{Data} & \textbf{per.} & \textbf{email} & \textbf{ph.} & \textbf{addr.} & \textbf{acct} & \textbf{date} \\
\midrule
ai4priv  & 0.777 & \textbf{0.969} & 0.933 & 0.853 & 0.958 & 0.671 \\
gretel   & 0.795 & 0.659 & 0.589 & 0.561 & 0.389 & 0.372 \\
kiji\_en & 0.394 & \textbf{0.975} & 0.920 & 0.304 & 0.856 & 0.533 \\
spy\_med & 0.406 & 0.504 & 0.485 & 0.456 & 0.672 & — \\
spy\_leg & 0.453 & 0.459 & 0.441 & 0.411 & 0.577 & — \\
\midrule
\textbf{Avg} & \textbf{.398} & \textbf{.782} & \textbf{.758} & \textbf{.491} & \textbf{.586} & \textbf{.547} \\
\bottomrule
\end{tabular}
\caption{Per-class span F1 (per.=person, ph.=phone, addr.=address, acct=account\#).
         Email and phone are near-perfect; person names and addresses fail most.}
\label{tab:perclass}
\end{table}

A consistent pattern emerges across domains: OPF is strongest on
\textbf{structurally regular} PII types (email: avg 0.78, phone: 0.76,
account numbers: 0.59) and weakest on \textbf{culturally variable} entities
(person names: 0.40, addresses: 0.49).
Email addresses and phone numbers share a near-universal format across
languages, making them easy to detect regardless of locale.
Person names and postal addresses are highly culture- and language-specific,
explaining their lower performance especially in non-English settings
(kiji\_en person: 0.394, address: 0.304).

\paragraph{Precision-recall profile across domains.}
Table~\ref{tab:pr} shows OPF is recall-biased across most PII domains (P=0.31--0.54, R=0.70--0.85) --- a model-level tendency rather than domain-specific behavior.
Exceptions: financial text is precision-biased (gretel\_en: P=0.64, R=0.49); general NER shows low recall (multiconer\_en: P=0.68, R=0.28) due to domain shift.
The gretel\_en exception is structurally explained: formatted identifiers (account numbers, IBANs, credit cards) are detected with high precision, but narrative financial PII (employer names, benefit descriptions) is missed, depressing recall.
Engineers should expect false-positive rates of 40--55\% in non-financial domains.

\begin{table}[h]
\centering
\small
\setlength{\tabcolsep}{5pt}
\begin{tabular}{lrrr}
\toprule
\textbf{Dataset} & \textbf{Precision} & \textbf{Recall} & \textbf{F1} \\
\midrule
ai4privacy          & 0.862 & 0.849 & 0.855 \\
nemotron            & 0.464 & 0.704 & 0.559 \\
kiji\_en            & 0.462 & 0.841 & 0.596 \\
kiji\_es            & 0.538 & 0.805 & 0.645 \\
gretel\_en          & 0.638 & 0.490 & 0.554 \\
spy\_medical        & 0.324 & 0.818 & 0.464 \\
spy\_legal          & 0.314 & 0.814 & 0.453 \\
conll2003\_en       & 0.608 & 0.534 & 0.569 \\
multiconer\_en      & 0.677 & 0.281 & 0.397 \\
\bottomrule
\end{tabular}
\caption{OPF precision, recall, F1 by dataset. OPF is recall-biased across
         most PII domains (prioritizes coverage over precision); gretel\_en is a notable
         exception (precision-biased). NER benchmarks show low recall due to domain shift.}
\label{tab:pr}
\end{table}

\section{Discussion}

\paragraph{OPF vs.\ Presidio: the production deployment question.}
OPF outperforms Presidio on all PII-annotated datasets (F1 gains: +0.42 AI4Privacy, +0.12 Nemotron, +0.17 Gretel/Kiji, +0.19/+0.17 SPY medical/legal) and also on CoNLL-2003 newswire NER (+0.18).
On MultiCoNER general NER, Presidio (0.497) surpasses OPF (0.397) because its spaCy PERSON recognizer covers the NER class without domain adaptation.
OPF is the stronger zero-shot choice for PII detection; Presidio retains an edge only on general NER.

\paragraph{GPT-4o as a zero-shot ceiling.}
Evaluated on six English PII datasets at $n$=2000 (Kiji $n$=846), GPT-4o outperforms OPF on
three: SPY medical (0.702 vs.\ 0.464), SPY legal (0.587 vs.\ 0.453),
and Gretel financial (0.576 vs.\ 0.554).
OPF leads on structured synthetic PII where span boundaries are regular:
AI4Privacy (0.855 vs.\ 0.543), Kiji customer support (0.596 vs.\ 0.420),
and Nemotron (0.559 vs.\ 0.478).
The pattern is consistent: GPT-4o's recall-heavy extraction
(SPY medical R\,=\,0.949 vs.\ OPF 0.818) recovers PII that OPF's
eight-category privacy schema misses, but its generative span recovery misses
exact boundaries on highly structured synthetic PII.
We do not ablate the mechanism; the pattern is reported as an empirical finding.

\paragraph{GLiNER as a zero-shot alternative.}
GLiNER~\cite{zaratiana-etal-2024-gliner} trails OPF on PII (AI4Privacy: 0.577 vs.\ 0.855; Kiji: 0.349 vs.\ 0.596) but is competitive on SPY medical (0.498 vs.\ 0.464) and dominates on general NER (MultiCoNER: 0.754 vs.\ 0.397; CoNLL: 0.428 vs.\ OPF 0.569).
GLiNER is preferable when the entity taxonomy is open or domain-specific; OPF is the stronger zero-shot choice for standard PII.

\paragraph{The labeled-data crossover.}
Per-class fine-tuning reaches 0.921 at $n$=5000, still 0.040 below binary's ceiling at full data.
Practitioners should adopt per-class labels only when ${\geq}$2000 annotated examples are available.

\paragraph{Inference latency and deployment profile.}
We did not measure throughput; the following ordering is qualitative, based on architectural characteristics: Presidio (CPU rule-based) $\gg$ XLM-R (340M, GPU-optional) $\approx$ GLiNER $>$ OPF (1.5B, MoE + Viterbi CRF, GPU required) $\gg$ GPT-4o (per-call API latency).
OPF's advantage over Presidio concentrates on culturally variable entities (PERSON, ADDRESS) where Presidio's spaCy NER struggles.

\section{Conclusion}

We presented a comprehensive evaluation of OpenAI's Privacy Filter across
32 datasets, 14 languages, and 5 domains.
OPF achieves strong zero-shot PII detection performance --- F1\,=\,0.46 on
synthetic medical/legal text (SPY), 0.60 on customer support (Kiji), and
0.86 on synthetic PII (AI4Privacy) --- substantially exceeding both
Microsoft Presidio (zero-shot) and XLM-RoBERTa-large-NER (domain-transferred, not fine-tuned on target datasets) across all PII domains.
However, performance degrades sharply outside OPF's training distribution:
on general NER benchmarks and especially for non-Latin scripts.
Learning curve experiments reveal a domain-dependent crossover:
$\sim$500 labeled examples on English synthetic PII ($\sim$100 on non-English Kiji), $\sim$1000 on synthetic medical PII;
per-class fine-tuning is less data-efficient than binary at small $n$,
validating binary labels as the more practical low-resource choice.
Error analysis reveals that structural regularity of PII type is the primary
predictor of detection performance.

We release all evaluation code, dataset converters, and results at:
\href{https://anonymous.4open.science/r/openai-privacy-filter-evaluation-48FF}{anonymous code repository}.

\section*{Limitations}

\paragraph{OPF fine-tuning.}
We evaluate OPF exclusively in zero-shot mode.
OPF supports fine-tuning via its documented training interface~\cite{opf2026}.
We evaluate exclusively in zero-shot mode to assess out-of-the-box performance.
The decision heuristic (Figure~\ref{fig:guide}) therefore reflects zero-shot OPF:
if OPF is fine-tuned on the target domain, the crossover point
($\sim$500 examples for English synthetic PII, $\sim$1000 for medical) would likely shift,
and XLM-RoBERTa may not be the optimal choice above that threshold.
Evaluating fine-tuned OPF is a natural extension of this work.

\paragraph{Learning curve generalization.}
The English crossover is established on three datasets: AI4Privacy and Gretel Financial
(synthetic) and SPY medical (synthetic).
Non-English curves are established on Kiji customer-support PII only (fr/de/es, binary labels);
whether per-class fine-tuning behaves similarly to the English finding (binary more efficient
at small $n$) is untested for non-English and is a natural next step.
Learning curves for SPY legal, Nemotron, and non-Latin-script datasets remain future work.
The domain-dependent crossover pattern (English synthetic: $\sim$500 examples; non-English Kiji: $\sim$100 examples;
medical: $\sim$1000 examples) should be treated as dataset-specific, not universally valid.

\paragraph{Script-family confounding.}
The non-Latin script analysis uses MultiCoNER PER-only spans evaluated against OPF's
\texttt{private\_person} class, so entity-type mismatch is not a confound.
However, script effects and OPF training data coverage cannot be disentangled:
low Arabic/Cyrillic F1 may reflect insufficient training examples in those scripts
rather than tokenizer fragmentation alone.
We report the gap as a descriptive finding; the fertility--F1 correlation is suggestive but not causal.
A clean test requires a multilingual PII dataset with Arabic and Cyrillic annotations,
which does not currently exist at scale.

\paragraph{XLM-RoBERTa label scheme.}
Most experiments use binary B-PII/I-PII/O labels for cross-dataset comparability.
Per-class fine-tuning (Section~\ref{sec:curves}) shows binary labels are more
data-efficient at small $n$, validating this choice.
Per-class fine-tuning against OPF's full 33-class output schema remains future work.

\paragraph{Presidio multilingual coverage.}
Presidio's multilingual support is limited to languages with available spaCy NER
models; in our experiments, direct language-matched evaluation is available for
English, French, German, and Spanish only (Table~\ref{tab:main_multi}).
For Danish, Dutch, and non-Latin scripts, Presidio falls back to English models
or lacks coverage entirely.
This understates Presidio's performance for languages it does support natively
and overstates it for languages it does not; the OPF vs.\ Presidio comparison
should be interpreted language-by-language rather than as a global claim.

\section*{Ethics Statement}

All datasets used are publicly available and contain either synthetic PII
or fully anonymized text.
No real personal data was collected or processed in this work.
Our evaluation reveals failure modes of OPF that practitioners should be
aware of before deployment in production PII-processing pipelines.

\bibliography{references}

@misc{opf2026,
  title        = {{OpenAI Privacy Filter}},
  author       = {{OpenAI}},
  year         = {2026},
  howpublished = {\url{https://github.com/openai/privacy-filter}},
}

@inproceedings{conneau-etal-2020-unsupervised,
  title     = {Unsupervised Cross-lingual Representation Learning at Scale},
  author    = {Conneau, Alexis and Khandelwal, Kartikay and Goyal, Naman and
               Chaudhary, Vishrav and Wenzek, Guillaume and Guzm{\'a}n, Francisco
               and Grave, Edouard and Ott, Myle and Zettlemoyer, Luke and
               Stoyanov, Veselin},
  booktitle = {Proceedings of ACL},
  year      = {2020},
  pages     = {8440--8451},
}

@inproceedings{multiconer2023,
  title     = {{MultiCoNER} v2: a Large Multilingual Dataset for Fine-Grained and Noisy Named Entity Recognition},
  author    = {Fetahu, Besnik and Chen, Zhiyu and Kar, Sudipta and Rokhlenko, Oleg and Malmasi, Shervin},
  booktitle = {Findings of the Association for Computational Linguistics: EMNLP 2023},
  year      = {2023},
  pages     = {2027--2051},
}

@misc{ai4privacy,
  title        = {{AI4Privacy PII Masking Dataset}},
  author       = {{AI4Privacy}},
  year         = {2023},
  howpublished = {\url{https://huggingface.co/datasets/ai4privacy/pii-masking-300k}},
}

@misc{nemotron,
  title        = {{Nemotron-PII}: Synthesized Data for Privacy-Preserving {AI}},
  author       = {Steier, Amy and Manoel, Andre and Haushalter, Alexa and {Van Segbroeck}, Maarten},
  year         = {2025},
  howpublished = {\url{https://huggingface.co/datasets/nvidia/Nemotron-PII}},
}

@misc{gretel,
  title        = {{Synthetic-PII-Financial-Documents-North-America}: A Synthetic Dataset for Training Language Models to Label and Detect {PII} in Domain Specific Formats},
  author       = {Watson, Alex and Meyer, Yev and {Van Segbroeck}, Maarten and Grossman, Matthew and Torbey, Sami and Mlocek, Piotr and Greco, Johnny},
  year         = {2024},
  howpublished = {\url{https://huggingface.co/datasets/gretelai/synthetic_pii_finance_multilingual}},
}

@misc{kiji,
  title        = {{Kiji PII Training Data}},
  author       = {{Dataiku}},
  year         = {2024},
  howpublished = {\url{https://huggingface.co/datasets/DataikuNLP/kiji-pii-training-data}},
}

@inproceedings{spy2025,
  title     = {{SPY}: Enhancing Privacy with Synthetic {PII} Detection Dataset},
  author    = {Savkin, Maksim and Ionov, Timur and Konovalov, Vasily},
  booktitle = {Proceedings of NAACL Student Research Workshop},
  year      = {2025},
}

@inproceedings{conll2002,
  title     = {Introduction to the {CoNLL}-2002 Shared Task:
               Language-Independent Named Entity Recognition},
  author    = {Tjong Kim Sang, Erik F.},
  booktitle = {Proceedings of COLING-02 Workshop on Shared Tasks and Comparative Evaluation},
  year      = {2002},
}

@inproceedings{conll,
  title     = {Introduction to the {CoNLL}-2003 Shared Task:
               Language-Independent Named Entity Recognition},
  author    = {Tjong Kim Sang, Erik F. and De Meulder, Fien},
  booktitle = {Proceedings of CoNLL},
  year      = {2003},
  pages     = {142--147},
}

@inproceedings{lison-etal-2021-anonymisation,
  title     = {Anonymisation Models for Text Data: State of the Art, Challenges and Future Directions},
  author    = {Lison, Pierre and Pil{\'a}n, Ildik{\'o} and Sanchez, David and
               Batet, Montserrat and {\O}vrelid, Lilja},
  booktitle = {Proceedings of ACL-IJCNLP},
  year      = {2021},
  pages     = {4188--4203},
}

@article{pilan-etal-2022-text,
  title   = {The Text Anonymization Benchmark ({TAB}): A Dedicated Corpus and Evaluation Framework for Text Anonymization},
  author  = {Pil{\'a}n, Ildik{\'o} and Lison, Pierre and {\O}vrelid, Lilja and
             Papadopoulou, Anthi and S{\'a}nchez, David and Batet, Montserrat},
  journal = {Computational Linguistics},
  year    = {2022},
  pages   = {1053--1101},
}

@inproceedings{zaratiana-etal-2024-gliner,
  title     = {{GLiNER}: Generalist Model for Named Entity Recognition
               using Bidirectional Transformer},
  author    = {Zaratiana, Urchade and Tomeh, Nadi and
               Holat, Pierre and Charnois, Thierry},
  booktitle = {Proceedings of NAACL},
  year      = {2024},
}

@misc{presidio,
  title        = {{Microsoft Presidio: Data Protection and De-identification SDK}},
  author       = {{Microsoft}},
  year         = {2023},
  howpublished = {\url{https://github.com/microsoft/presidio}},
}

\end{document}